\documentclass{svproc}
\usepackage{amsmath,amssymb,amsfonts}
\usepackage{algorithmic}
\usepackage{graphicx}
\usepackage{textcomp}
\usepackage{xcolor}
\usepackage{float}
\usepackage[justification=centering]{subfig}
\usepackage{placeins}
\usepackage{ragged2e}

\let\svprocsubparagraph\subparagraph
\let\subparagraph\paragraph
\usepackage{titlesec}
\let\subparagraph\svprocsubparagraph

\usepackage{scalerel}
\usepackage[misc]{ifsym}
\usepackage{tikz}
\usetikzlibrary{svg.path}
\usepackage{fancyhdr}
\usepackage{hyperref}
\hypersetup{
    colorlinks=false,
    citecolor=black,
    linkcolor=black,
    filecolor=black,
    urlcolor=blue
}

\definecolor{orcidlogocol}{HTML}{A6CE39}
\tikzset{
  orcidlogo/.pic={
    \fill[orcidlogocol] svg{M256,128c0,70.7-57.3,128-128,128C57.3,256,0,198.7,0,128C0,57.3,57.3,0,128,0C198.7,0,256,57.3,256,128z};
    \fill[white] svg{M86.3,186.2H70.9V79.1h15.4v48.4V186.2z}
                 svg{M108.9,79.1h41.6c39.6,0,57,28.3,57,53.6c0,27.5-21.5,53.6-56.8,53.6h-41.8V79.1z M124.3,172.4h24.5c34.9,0,42.9-26.5,42.9-39.7c0-21.5-13.7-39.7-43.7-39.7h-23.7V172.4z}
                 svg{M88.7,56.8c0,5.5-4.5,10.1-10.1,10.1c-5.6,0-10.1-4.6-10.1-10.1c0-5.6,4.5-10.1,10.1-10.1C84.2,46.7,88.7,51.3,88.7,56.8z};
  }
}

\newcommand\orcidicon[1]{\href{https://orcid.org/#1}{\mbox{\scalerel*{
\begin{tikzpicture}[yscale=-1,transform shape]
\pic{orcidlogo};
\end{tikzpicture}
}{|}}}}

\usepackage{hyperref}

\titlespacing\section{0pt}{12pt plus 4pt minus 2pt}{0pt plus 2pt minus 2pt}
\titlespacing\subsection{0pt}{12pt plus 4pt minus 2pt}{0pt plus 2pt minus 2pt}
\titlespacing\subsubsection{0pt}{12pt plus 4pt minus 2pt}{0pt plus 2pt minus 2pt}
\def \hfillx {\hspace*{-\textwidth} \hfill}
\usepackage{url}

\begin{document}

\mainmatter              % start of a contribution
\title{A Tiny CNN Architecture for Medical Face Mask Detection for Resource-Constrained Endpoints}
\titlerunning{A tiny CNN architecture for medical face mask detection}  % abbreviated title (for running head)
\author{Puranjay Mohan \orcidicon{0000-0002-8835-7586}\inst{1, 2}\textsuperscript{(\Letter)} \and Aditya Jyoti Paul \orcidicon{0000-0002-4351-2108}\inst{1, 3} \and Abhay Chirania \orcidicon{0000-0002-7809-8779}\inst{2}}
\authorrunning{Puranjay Mohan et al.} % abbreviated author list (for running head)
\tocauthor{Puranjay Mohan, Aditya Jyoti Paul, Abhay Chirania}
\institute{Cognitive Applications Research Lab, India\\
\and
Department of Electronics and Communication Engineering,\\SRM Institute of Science and Technology, Kattankulathur,\\Tamil Nadu – 603203, India\\
\and
Department of Computer Science and Engineering,\\SRM Institute of Science and Technology, Kattankulathur,\\Tamil Nadu – 603203, India\\
\email{\{puranjaymohan\_mu, aditya\_jyoti, abhaychirania\_de\}@srmuniv.edu.in}
}
\maketitle              % typeset the title of the contribution
\vspace{-5mm}
\thispagestyle{fancy} 
\fancyhf{}
\renewcommand{\headrulewidth}{0pt}
\lfoot{\textit{\footnotesize This is a post-print of a paper published in Innovations in Electrical and Electronic Engineering. Lecture Notes in Electrical Engineering, vol 756. Springer, Singapore.  \href{https://link.springer.com/chapter/10.1007\%2F978-981-16-0749-3_52}
{\color{blue} doi: 10.1007/978-981-16-0749-3\_52.} \href{https://www.springer.com/series/7818}}}
\begin{abstract}
The world is going through one of the most dangerous pandemics of all time with the rapid spread of the novel coronavirus (COVID-19). According to the World Health Organisation, the most effective way to thwart the transmission of coronavirus is to wear medical face masks. \\Monitoring the use of face masks in public places has been a challenge because manual monitoring could be unsafe. This paper proposes an architecture for detecting medical face masks for deployment on resource-constrained endpoints having extremely low memory footprints. A small development board with an ARM Cortex-M7 microcontroller clocked at480 Mhz and having just 496 KB of framebuffer RAM, has been used for the deployment of the model. Using the TensorFlow Lite framework, the model is quantized to further reduce its size. The proposed model is 138 KB post quantization and runs at the inference speed of 30 FPS.

\keywords{ARM Cortex-M7, COVID-19, Edge Computing, Face mask detection, Quantization, SqueezeNet, TinyML}
\end{abstract}

\section{Introduction}
The sudden increase of computational capability and availability of data in the last few years has allowed intelligent systems to solve various problems involving computer vision, speech, etc. Traditionally, these models got deployed on servers with high compute and storage capabilities. With the rise of the Internet of things and edge computing, the need to deploy these systems at the edge has grown. The major roadblock in edge deployment of deep neural networks is the very high computational and memory footprint of these models. Image classification is one such problem where edge deployment is in high demand because of the many applications which rely on it. Face mask detection is a subset of image classification where the goal is to classify the image into two classes, i.e. Mask and No-Mask, respectively.

The outbreak of the novel coronavirus has significantly impacted the livelihood of people across the globe \cite{covidEnc} and effective deployment of face mask detection in public places can help in thwarting the transmission of the virus. Face mask detection is a non-trivial problem because of its high throughput, reliability, and privacy requirements; it can't use traditional deployment methods where the image is first sent to a server for classification and the result is sent back for further use in the application.

One application scenario the authors of this research work envision is a small camera attached to an automatic door, the camera continuously takes images of the person standing in front of it and only opens if the person is wearing a mask. This intelligent door can be used at all public places and it will safely monitor the people entering public premises. One requirement of this smart door is that it should be cheap and consume less energy. The proposed model is small enough to fit inside the memories of the smallest and cheapest microcontrollers available in the market. Many other applications requiring high speed mask-detection on the edge can be envisaged as other possible applications as well.

TinyML is an upcoming field at the intersection of hardware, software, and machine learning algorithms, that is gaining massive traction. Recent developments in this field include building deep neural networks having sizes of few hundreds of KBs. This paper presents the process to train and deploy an innovative architecture on the OpenMV H7 development board for detecting face masks using the small on-board camera.

A major challenge with TinyML is that most microcontrollers don't have a floating point unit and hence all mathematical computations need to work on integers. This leads to a smaller model along with a change in accuracy that is difficult to predict. Earlier studies focused on deployment on more powerful edge devices, using much larger models, but this paper reports how the quantized CNN model compares to existing architectures in terms of model size, accuracy and performance. 

The rest of the paper is organised as follows - Section 2 covers the literature review, Section 3 discusses some technical and hardware details, Section 4 explains the experimental methodology, Section 5 reports the observations and the findings, and section 6 concludes this paper, throwing some light on possible future research avenues in this field.

\section{Literature Review}
In this section, some prior advances in face mask detection and quantization, which are the primary facets of this research work, have been reviewed.
\vspace{-2mm}
\subsection{Face Mask Detection}
Due to the Coronavirus pandemic, Face Masks have become an integral part of our society, hence numerous implementations of detecting face mask have come forward which are based on Convolutional Neural Networks \cite{Lecun98gradient-basedlearning}. \cite{chavda2020multistage} proposed a two stage detection scheme, the first being face detection, and the next being face mask classifier. \cite{99a5a804e1134f46a93a858fb3cb597b} proposed a hybrid deep transfer learning model with two components, the first for feature extraction using ResNet50 and other for classification using SVM, and other ensemble algorithms. Their Support Vector Machine (SVM) classifier achieved testing accuracy of 99.64\%. RetinaMask \cite{jiang2020retinamask} achieved state-of-the-art result on public face mask dataset (2.3\% and 1.5\% higher precision than baseline result on face and mask detection) using one-stage detector which consisted of feature pyramid network to fuse high level semantic information with multiple feature maps. \cite{8888092} presented a model with pre-trained weights of VGG-16 architecture for feature extraction and then a fully connected neural network (FCNN) to segment out faces present in an image, and detect face masks on them. The model showed great result in recognizing non-frontal faces. \cite{LOEY2020102600} proposed a model using YOLO-v2 with ResNet-50 which achieves higher average precision by using mean Intersection over Union (IoU). \cite{chowdary2020face} proposed a model based on transfer learning with InceptionV3, It outperformed recently proposed models by achieving testing accuracy of 100\% on Simulated Masked Face Dataset (SMFD). \cite{Roy2020} discussed the challenge of implementing object detection on edge devices, the paper compared various popular object detection algorithms like YOLO-v3, YOLO-v3tiny, Faster R-CNN, etc. to determine the most efficient algorithm for real time detection of face masks.
\subsection{Quantization}
Leveraging quantization techniques is necessary for  implementing CNNs on resource-constrained devices. \cite{banner2019posttraining} introduced 4-bit training quantization on both activation and weights, achieving accuracies, a few percent less than state-of-the-art baselines across CNNs. \cite{nahshan2020loss} proposed a method which quantizes layer parameters that improve accuracy over existing post-training quantization techniques. \cite{zhao2019improving} proposed an outlier channel splitting (OCS) based method to improve quantization performance without retraining. \cite{choukroun2019lowbit} discussed low bit quantization of neural networks by optimization of constrained Mean Squared Error(MSE) problems for performing hardware-aware quantization. \cite{jacob2017quantization} proposed a quantization scheme along with a co-designed training procedure. The paper concluded that inference using integer-only arithmetic performs better than floating-point arithmetic on typical ARM CPUs. \cite{gong2019differentiable} proposed Differentiable Soft Quantization (DSQ) to bridge the gap between the full-precision and low-bit networks. The hybrid compression model in \cite{8019465} uses four major modules, Approximation, Quantization, Pruning, and Coding, which provides 20-30x compression rate with negligible loss in accuracy. The research by \cite{dong2019hawq}, \cite{dong2019hawqv2}, and \cite{yao2020hawqv3} proposed mixed-precision quantization techniques, where more sensitive layers are at higher precision.
\vspace{-2mm}
\section{Technical Details of the problem}
This section explains the technical details related to the experimental setup including the hardware, the software, and the use of datasets.
\vspace{-2mm}
\subsection{Hardware setup for deployment}
The hardware used in this research for edge deployment is the OpenMV Cam H7 \cite{abdelkader2017openmv}, housing STMicroelectronics' STM32H743VI \cite{stmicro}, an ARM Cortex-M7 based 32-bit microcontroller and a small camera. The microcontroller has a clock speed of 480 Mhz, 1 MB SRAM for various applications and 2 MB of flash memory for non-volatile storage. The development board provides a MicroPython based operating system allowing easy deployment and on-device analytics of TF-Lite models.

The documentation of the board suggests keeping the model under 400KB, but during this study \cite{ourpaper} we found that the biggest model which can be fit successfully in memory is under 230KB. A larger model of size upto 1MB can be stored on the flash memory but for that the model has to converted into a FlatBuffer using STM32Cube.AI and the operating system has to be recompiled, which leads to the loss of the utility of the MicroPython.

The models were trained using Kaggle kernels with TPU (Tensor Processing Unit) acceleration enabled, 128 TPU elements per core with 8 such cores running in parallel, this made the training time extremely short.

\subsection{Dataset Construction}
Four datasets were used in this research work. The details of the datasets can be seen in Table \ref{tab1}. The first dataset from Kaggle \cite{dataset12k} has around 11,792 images taken on different backgrounds and cropped to the face region. The images of this dataset were merged and interpolation augmentation was applied using OpenCV's interpolation methods, INTER\_AREA, INTER\_CUBIC, INTER\_NEAREST, INTER\_LINEAR, and INTER\_LANCZOS4 to augment the images to 58,960. 

The second dataset was also from Kaggle \cite{dataset2} which had 440 images taken on noisy backgrounds, equally divided into mask and without mask images. It was augmented to 22,200 using standard augmentation followed by interpolation. 

\begin{figure}[h]
    \centering
    \vspace{-0.8\baselineskip}
    \includegraphics[width=0.5\linewidth]{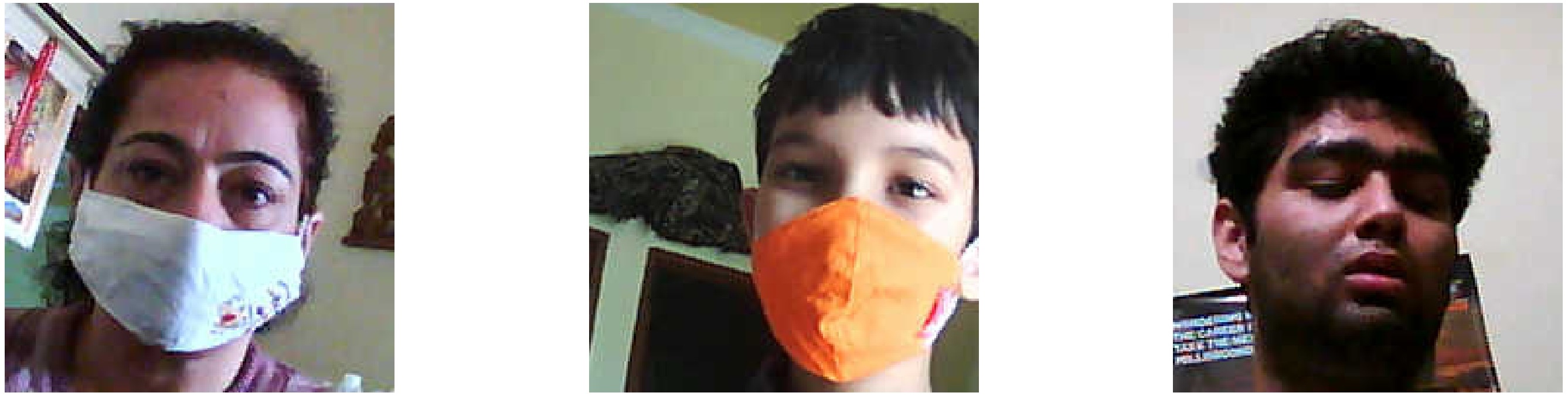}
    \vspace{-0.8\baselineskip}
    \caption{Images from OpenMV Cam H7 Dataset}
    \label{fig:dataset}
\end{figure}

The third dataset was produced by the authors using the OpenMV Cam H7 camera. Images of size 200x200 were taken and saved on the SD Card of the development board. This dataset had 1979 images which were augmented to 49,895 using augmentation techniques discussed earlier. Some images from this dataset can be seen in Fig. \ref{fig:dataset}.

A fourth dataset was also produced using the OpenMV camera which had 594 images augmented to 4794. This dataset was held out for testing the performance of the OpenMV Cam H7. The exact usage of this dataset is novel to this research work, and has been elaborated in Section 4.

All the images of each dataset were resized to 32x32 as it was found to be the optimal size of the image to fit in the framebuffer of the microcontroller.

\begin{table}[htbp]
\vspace{-2mm}
\caption{Dataset Details}
\begin{center}
\vspace{-0.8\baselineskip}
\scalebox{0.6}{
\begin{tabular}{c@{\quad}c@{\quad}c@{\quad}c}
\hline
\textbf{S.No} & \textbf{\textit{Name}}& \textbf{\textit{Original}}& \textbf{\textit{Augmented}} \\
\hline
1&Face Mask ~12K Images Dataset&11792&58960\\
\hline
2&Face Mask Classification&440&22200\\
\hline
3&OpenMV Datset&1979&49895\\
\hline
\multicolumn{2}{c}{\textbf{Total Data for training and validation}}&\textbf{14211}&\textbf{131055}\\
\hline
\multicolumn{2}{c}{\textbf{Total Data for testing}}&\textbf{594}&\textbf{4794}\\
\hline
\end{tabular}}
\label{tab1}
\end{center}
\end{table}
\vspace{2mm}
\vspace{2\baselineskip}
\section{Experimental Methodology}
This section explains the steps taken in this research work for data splitting, model design, evaluation, and model comparison.
\subsection{Data Splitting}
The datasets after being merged had 131,055 images in total, these included images from two Kaggle datasets\cite{dataset12k}\cite{dataset2} and one dataset produced by the OpenMV Camera. The fourth dataset was held out and was used for testing. 

This regime of holding out a separate dataset for testing is not usually performed but was considered imperative in this research work to evaluate the generalization achieved by the models running on the microcontroller. The normal regime of combining everything and then taking out a small portion for testing would be unable to show the true model performance on the target edge-case scenarios due to differences between the distributions of the train and test sets.

\subsection{Proposed Architecture and Comparison with SqueezeNet}

\begin{figure}[htbp]
    \centering
    \includegraphics[width=0.5\linewidth]{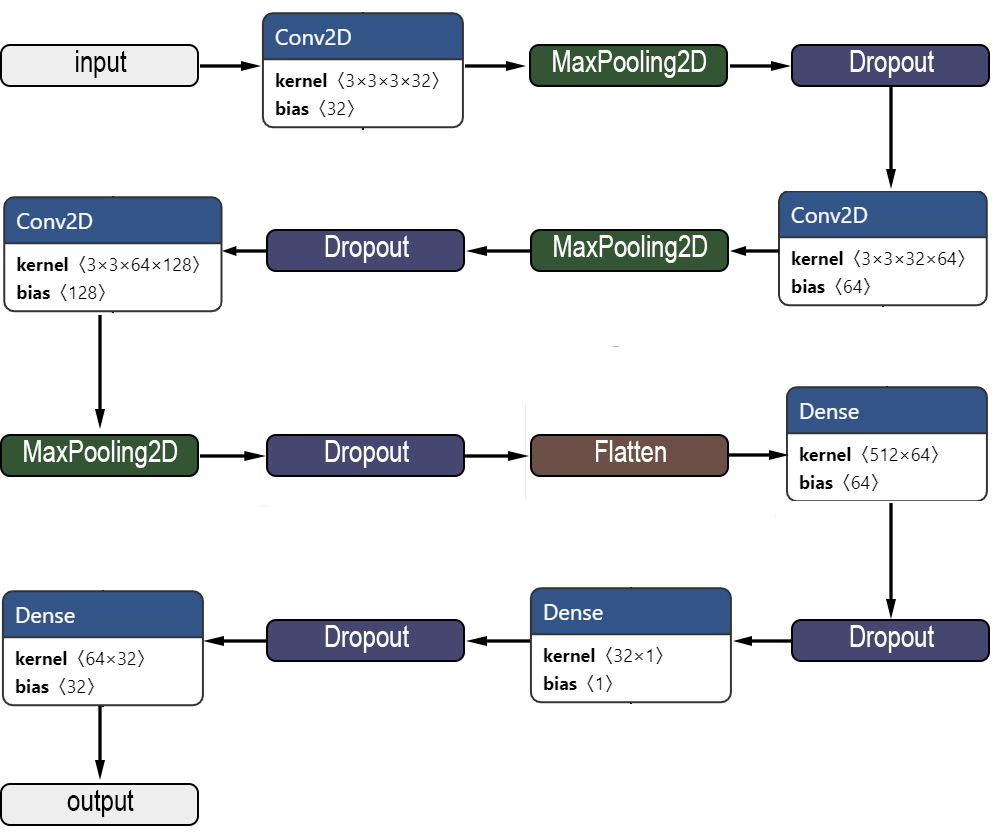}
    \caption{Proposed Architecture}
    \label{fig:model_image}
\end{figure}

After experimenting with different architectures and comparing their size and performance, the CNN architecture shown in Fig. \ref{fig:model_image} was found to be the best, after considering the RAM constraints of the device. This model has 128,193 trainable parameters and full integer quantization reduces it to 138 KB.

SqueezeNet\cite{iandola2016squeezenet} was chosen for comparison with the proposed model because of its small size. A smaller version of SqueezeNet was also built by removing two fire modules, which is called Modified SqueezeNet in this work, and also used for comparison.
\vspace{-1.5mm}
\subsection{Training Specifications}

All three models were designed in TensorFlow 2.3. BinaryCrossentropy loss was chosen as the loss function, shown in Eq. (\ref{eq}), because the problem involved binary classification of images. 
\vspace{-2mm}
\begin{equation}
Loss = -{(y\log(p) + (1 - y)\log(1 - p))}\label{eq}
\vspace{-2mm}
\end{equation}
\quad\quad\quad where $y$ is the true label and $p$ is the predicted label.

Adam was chosen as the optimizer with a learning rate of 0.001, first moment decay of 0.9, second moment decay of 0.999, and epsilon was chosen as 10\textsuperscript{-7}. ReduceLROnPlateau callback was used to reduce the learning rate by a factor of 0.2 when the validation accuracy didn't improve for 5 epochs. ModelCheckPoint callback was used to save the best weights into a file.

\subsection{Post-training Procedure}
TensorFlow-Lite's Full Integer Quantization was used to convert all three models from float-32 precision to Int-8 precision. This procedure used a representative dataset for this conversion using the dynamic range of activations. This representative dataset was built by taking a small part of the test set.

\subsection{Evaluation of the Quantized Models}
All three models were evaluated using the TensorFlow-Lite Interpreter. The quantized models were loaded into the Interpreter and OpenMV test set was used to find the classification metrics.

On-device evaluation was not possible for the SqueezeNet and the Modified SqueezeNet because both of them were bigger than 230 KB. The proposed model was loaded onto the OpenMV Cam, a script took images of size 200x200, scaled them to 32x32, and normalized them before feeding it to the model. All images and predictions were saved on the SD Card, and later used for the analysis.
\section{Results and Discussion}
This section explains the results and analysis, along with comparison of the proposed model against SqueezeNet and modified SqueezeNet.

\subsection{SqueezeNet Model}
The methodology discussed in sec. 4 was followed to train the SqueezeNet model. The training accuracy reached 99.79\%, achieving a test accuracy of 98.50\% for the float32 model and 98.53\% for the int8 model. The size of the float32 model was 8 MB which shrunk to 780 KB post quantization. The details can be seen in Fig. \ref{fig:squeez_acc}.
\vspace{-0.5\baselineskip}
\subsection{Modified SqueezeNet Model}
Modified SqueezeNet was also trained in a similar way. Size of the float32 model was 3.84 MB which got reduced to 386 KB after quantization. The test accuracies for this model were 98.93\% and 98.99\% for float32 and int8 respectively. The details can be seen in Fig. \ref{fig:modsqueez_acc}.

\subsection{Our Proposed Model}
\begin{figure}[H]
    \centering
    
    \subfloat[Loss Evolution]{
    \includegraphics[width=0.22\linewidth]{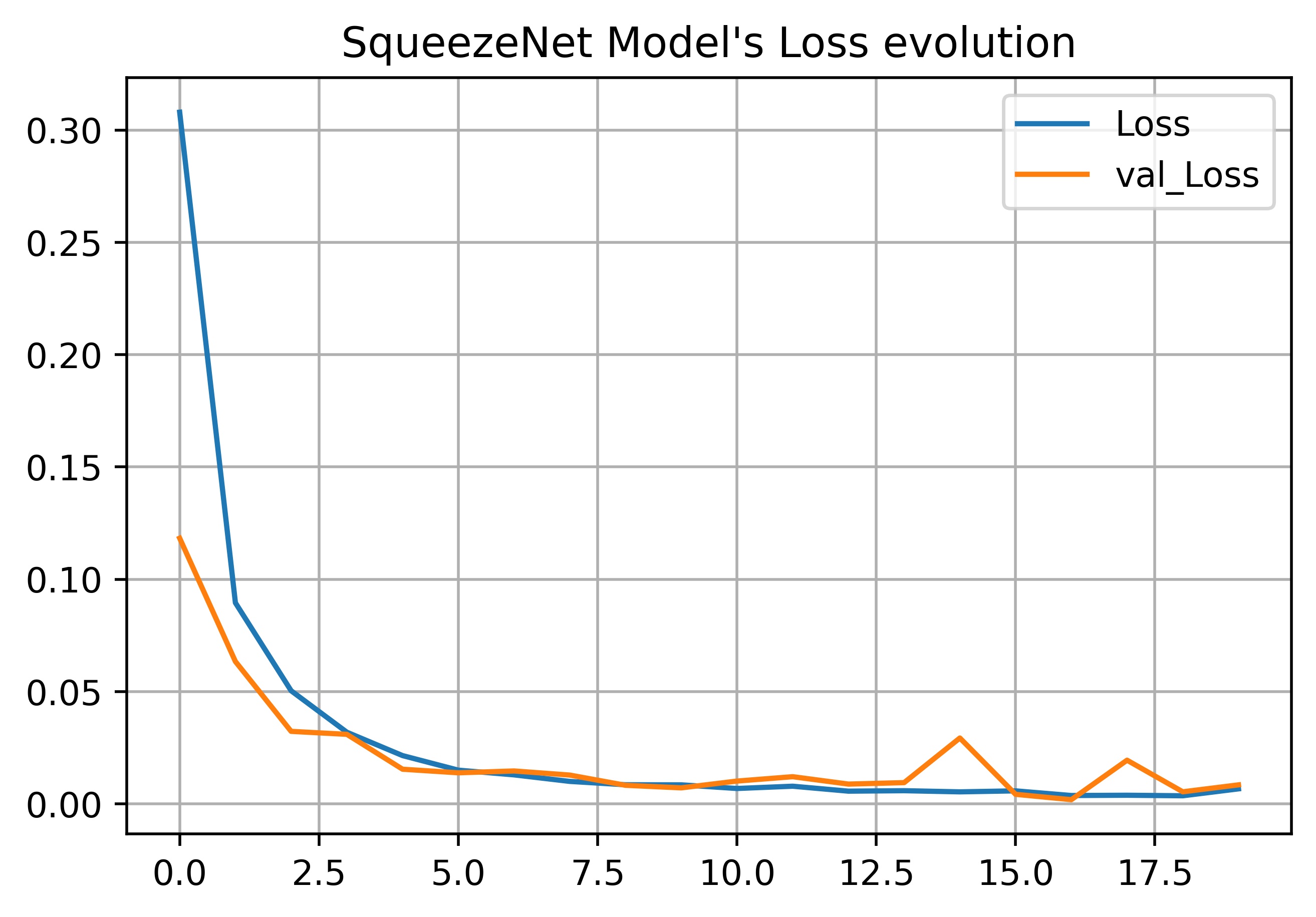}
    }
    \subfloat[  Accuracy\\  Evolution]{
    \includegraphics[width=0.24\linewidth]{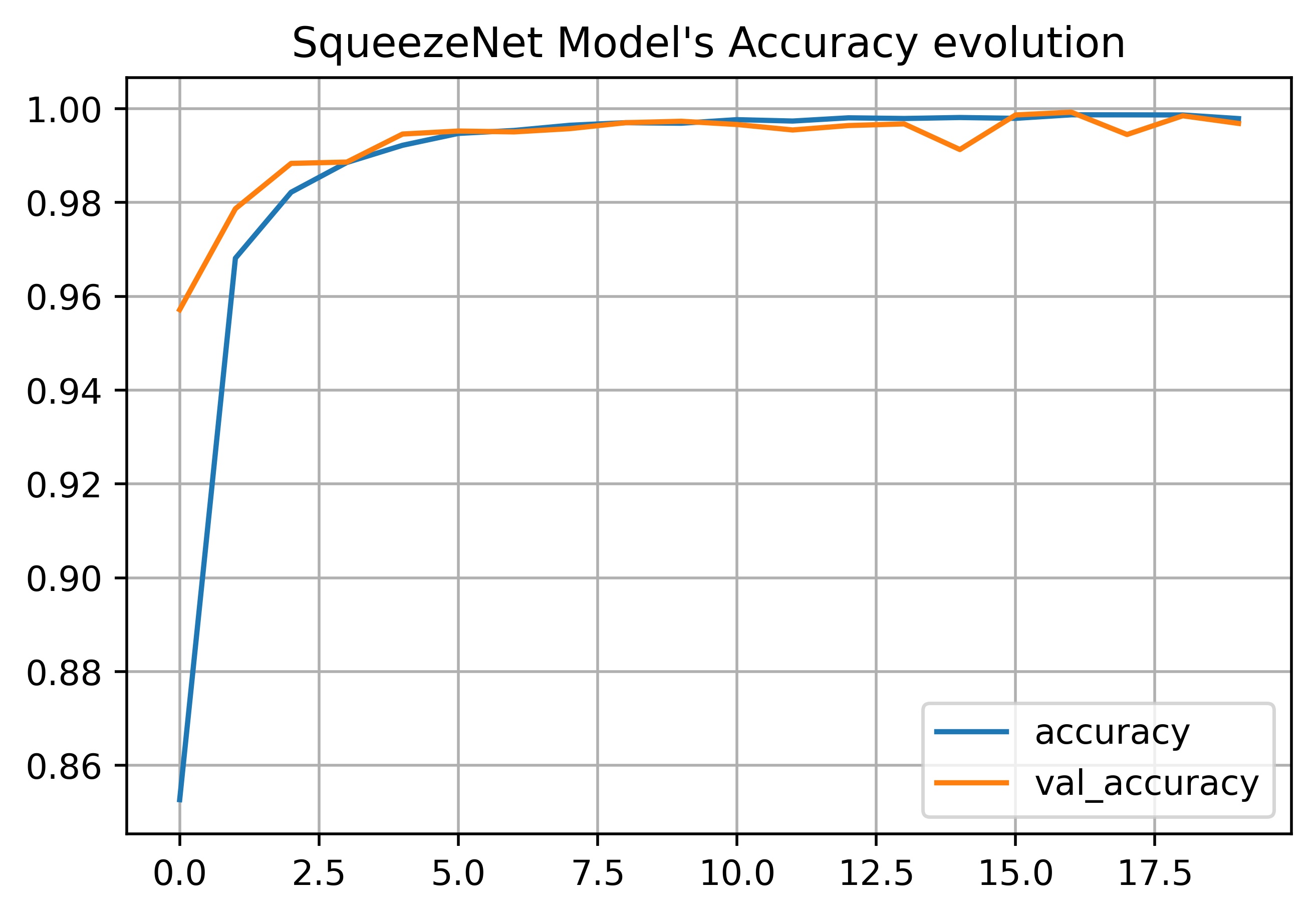}
    }
    \subfloat[Float32\\Confusion Matrix]{
    \includegraphics[width=0.24\linewidth]{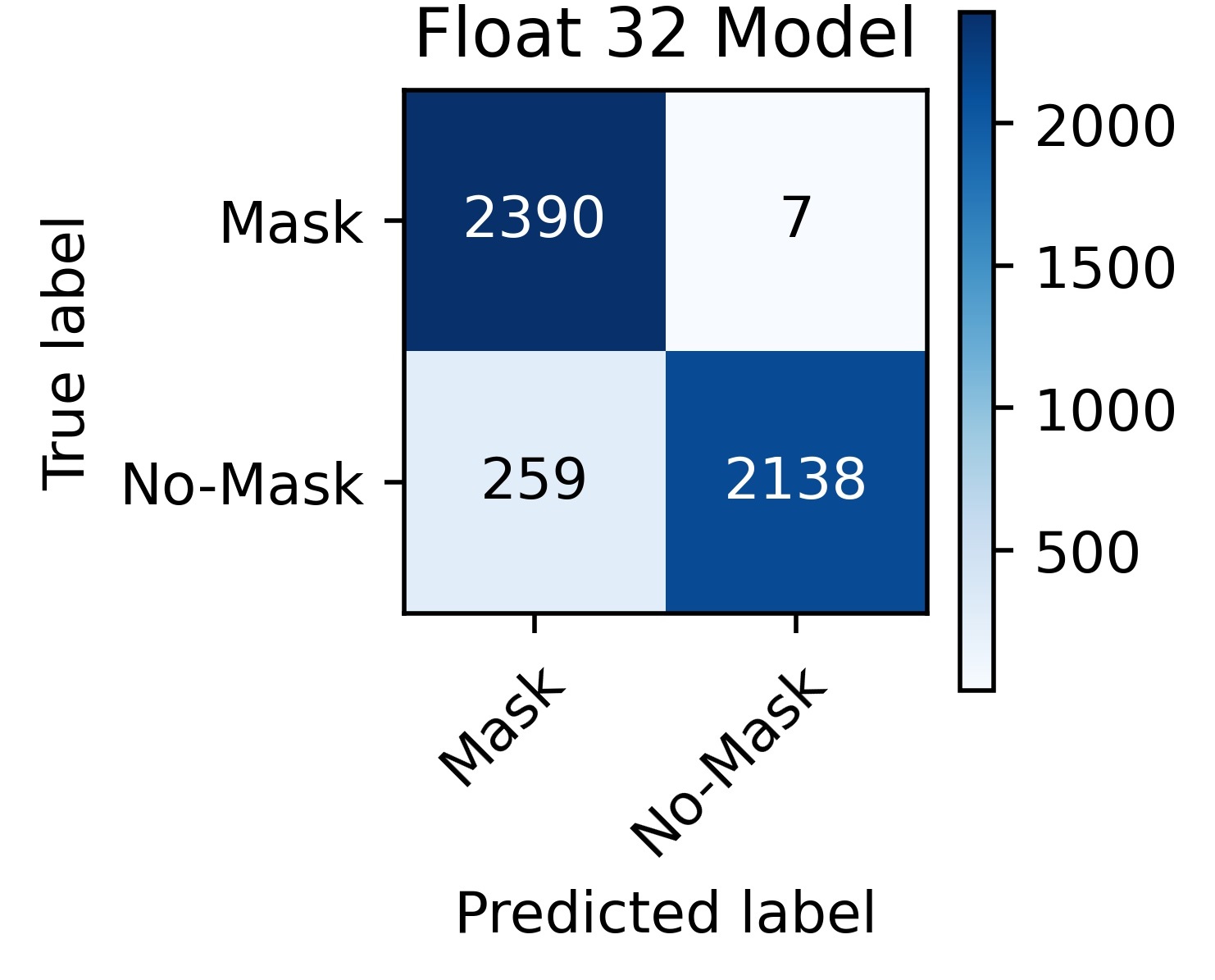}
    }
    \subfloat[int8\\Confusion Matrix]{
    \includegraphics[width=0.24\linewidth]{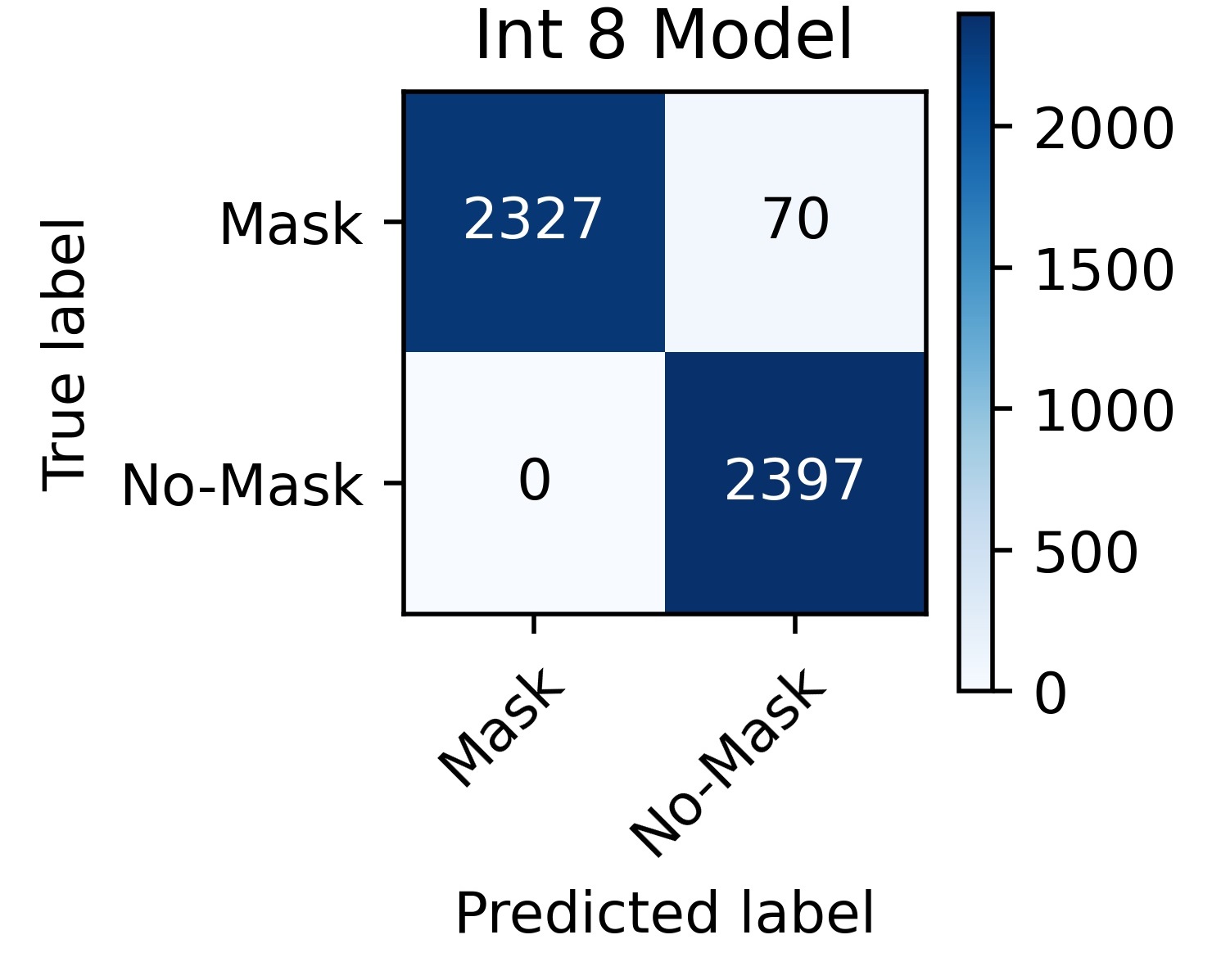}
    }\\
      \vspace{-2mm}
    \caption{SqueezeNet}
    \label{fig:squeez_acc}
    \vspace{-1mm}
    
    \centering
    \subfloat[Loss Evolution]{
    \includegraphics[width=0.22\linewidth]{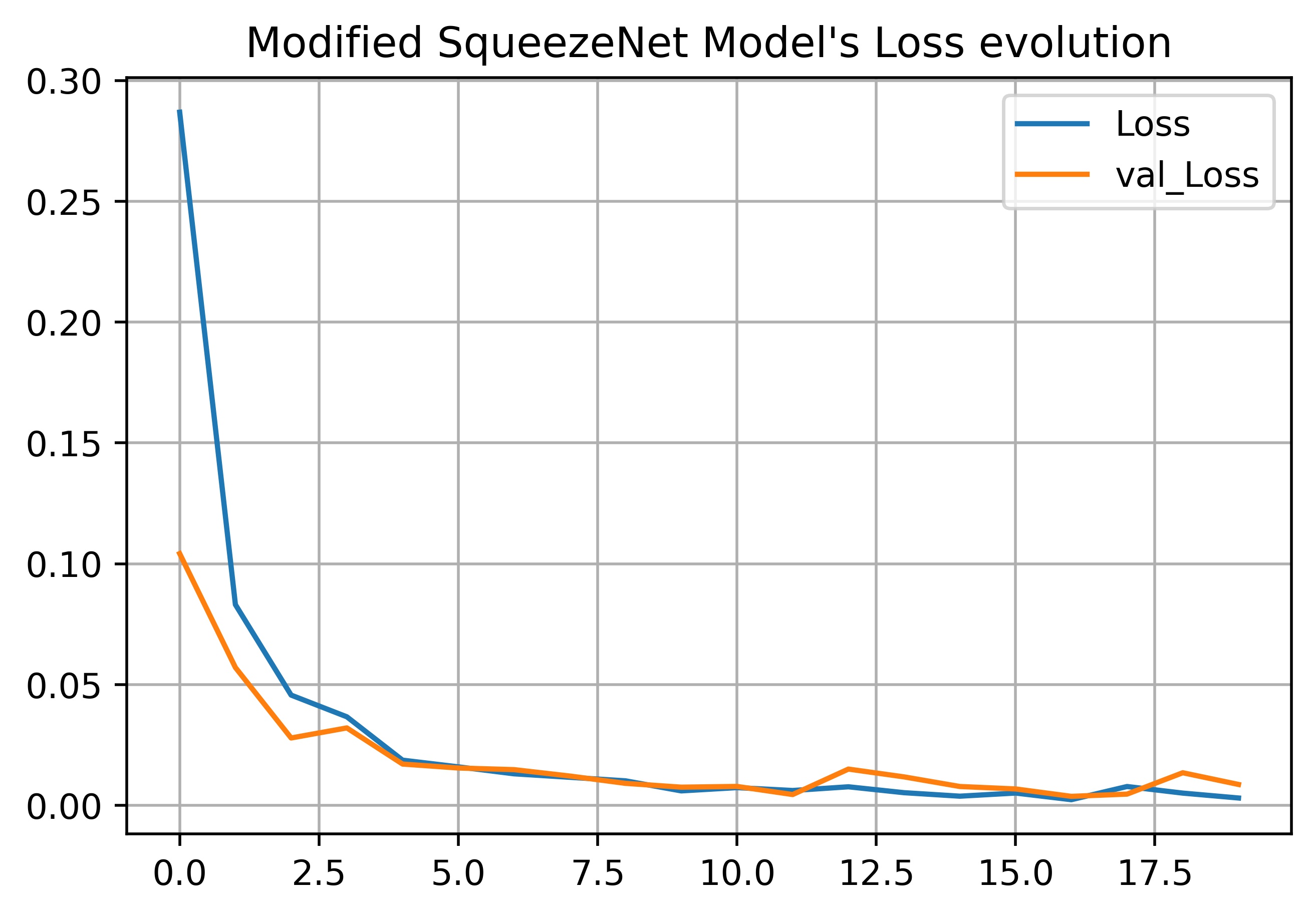}
    }
    \subfloat[Accuracy\\Evolution]{
    \includegraphics[width=0.24\linewidth]{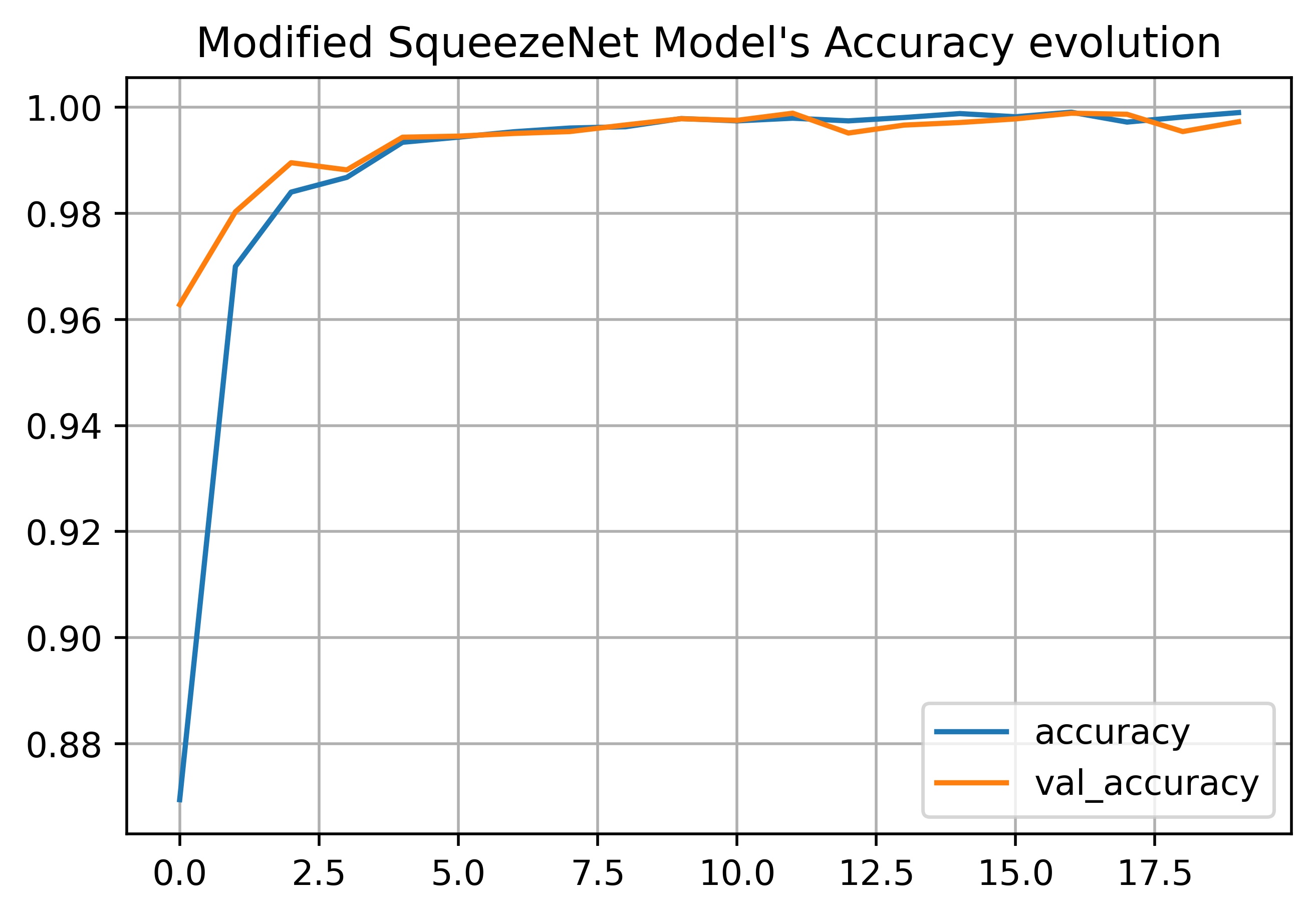}
    }
    \subfloat[Float32\\Confusion Matrix]{
    \includegraphics[width=0.24\linewidth]{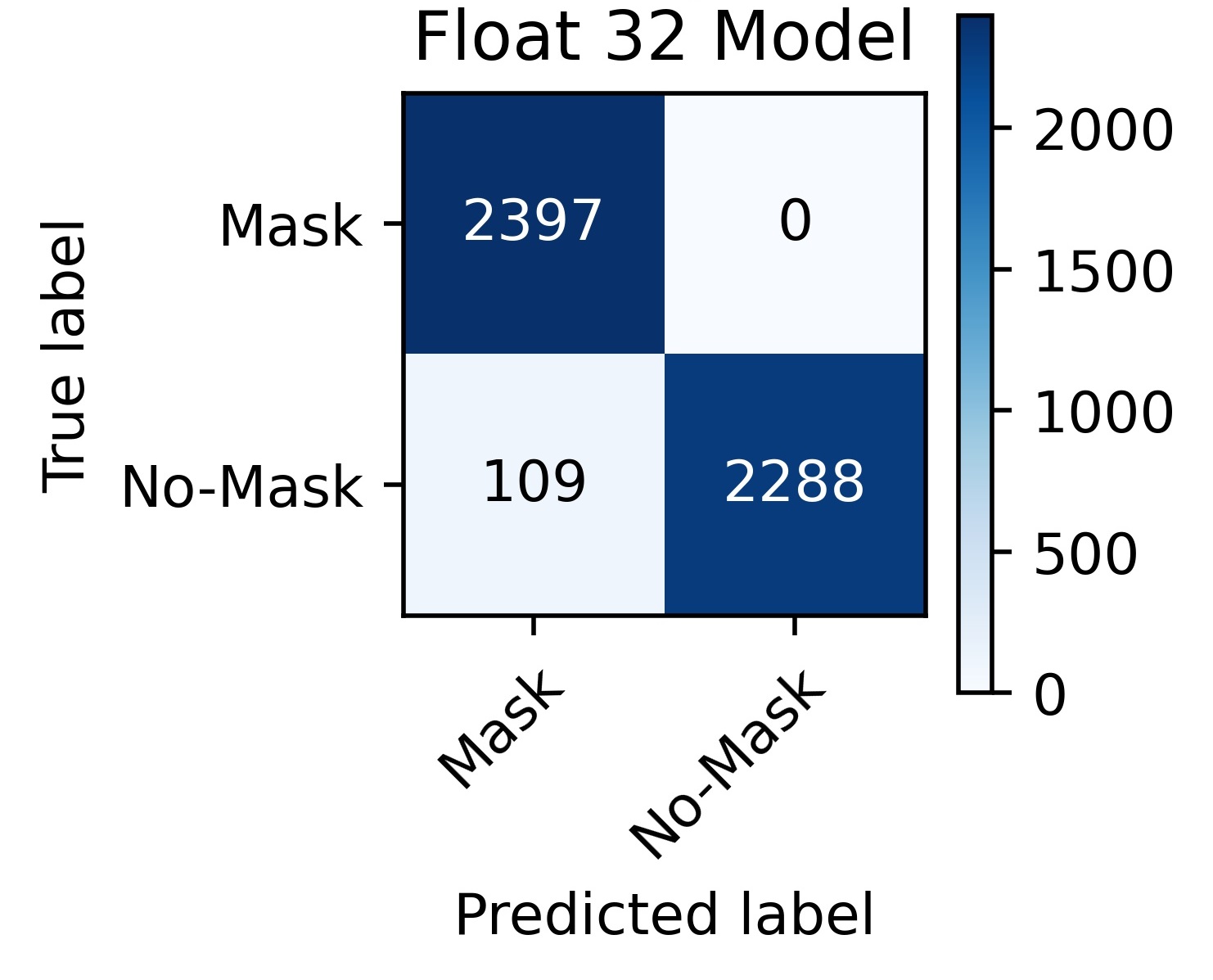}
    }
    \subfloat[Int8\\Confusion Matrix]{
    \includegraphics[width=0.24\linewidth]{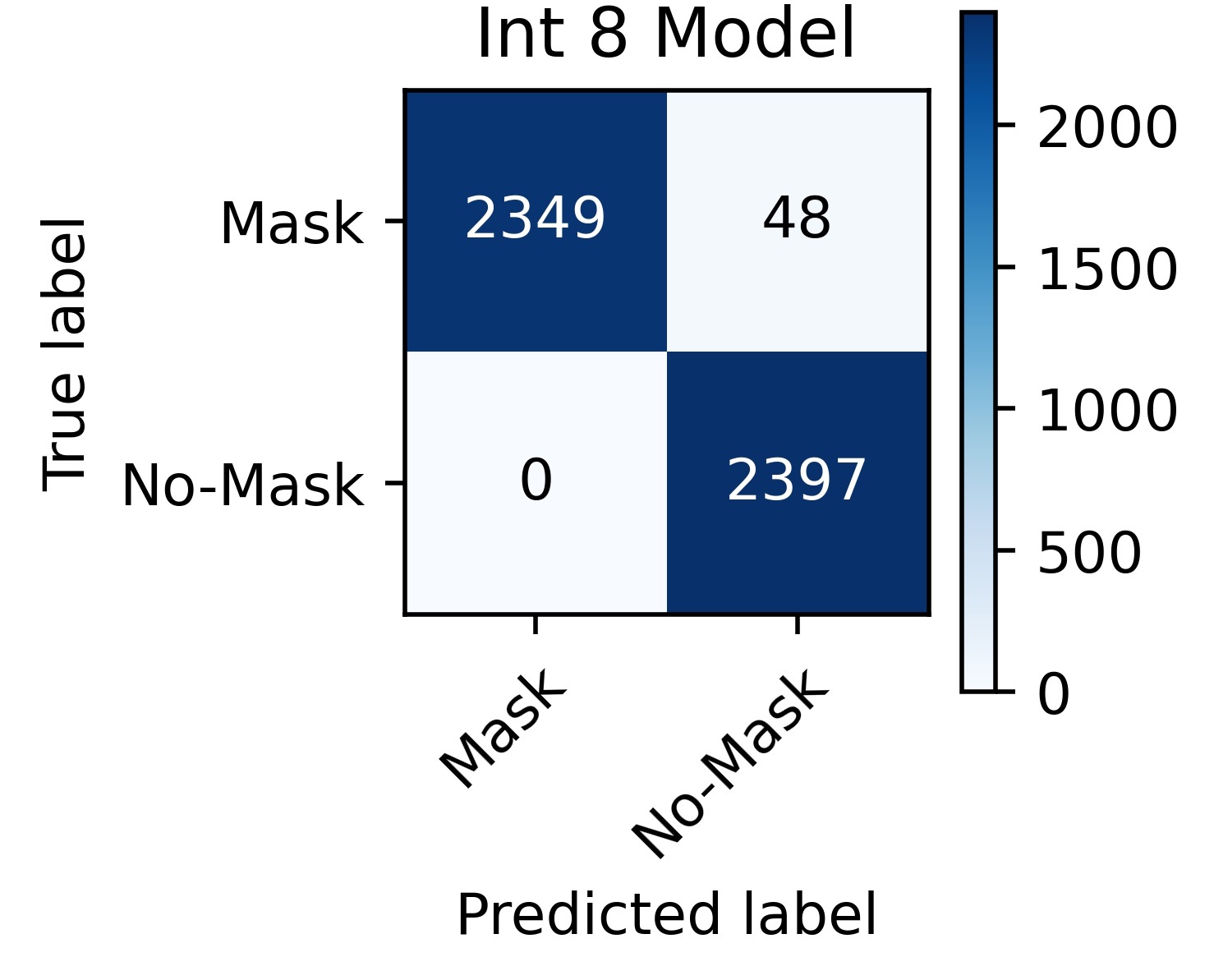}
    }\\
    \vspace{-1mm}
    \caption{Modified SqueezeNet}
    \label{fig:modsqueez_acc}
     \vspace{-2mm}

    \centering
    \subfloat[Loss Evolution]{
    \includegraphics[width=0.22\linewidth]{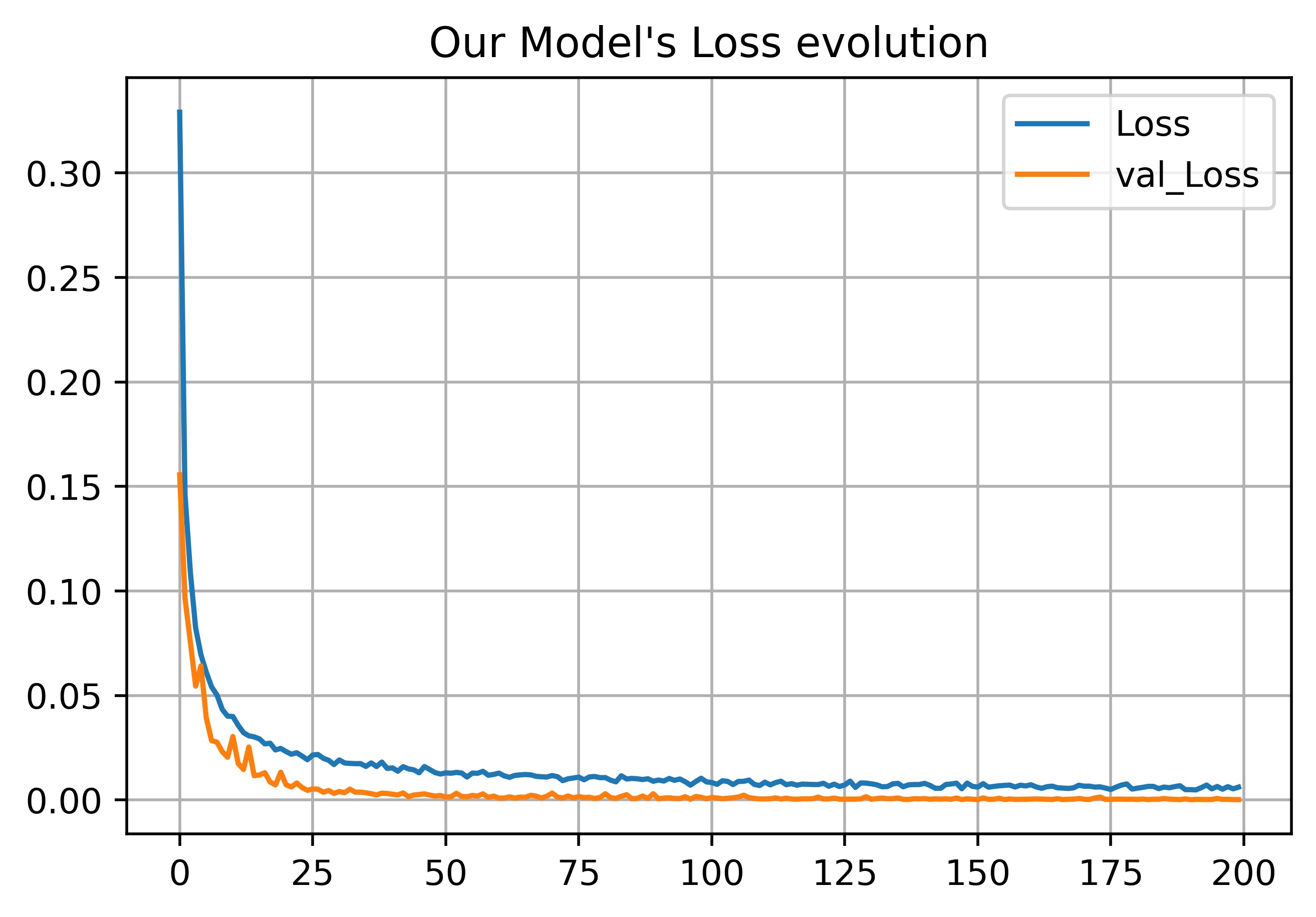}
    }
    \subfloat[Accuracy\\Evolution]{
    \includegraphics[width=0.24\linewidth]{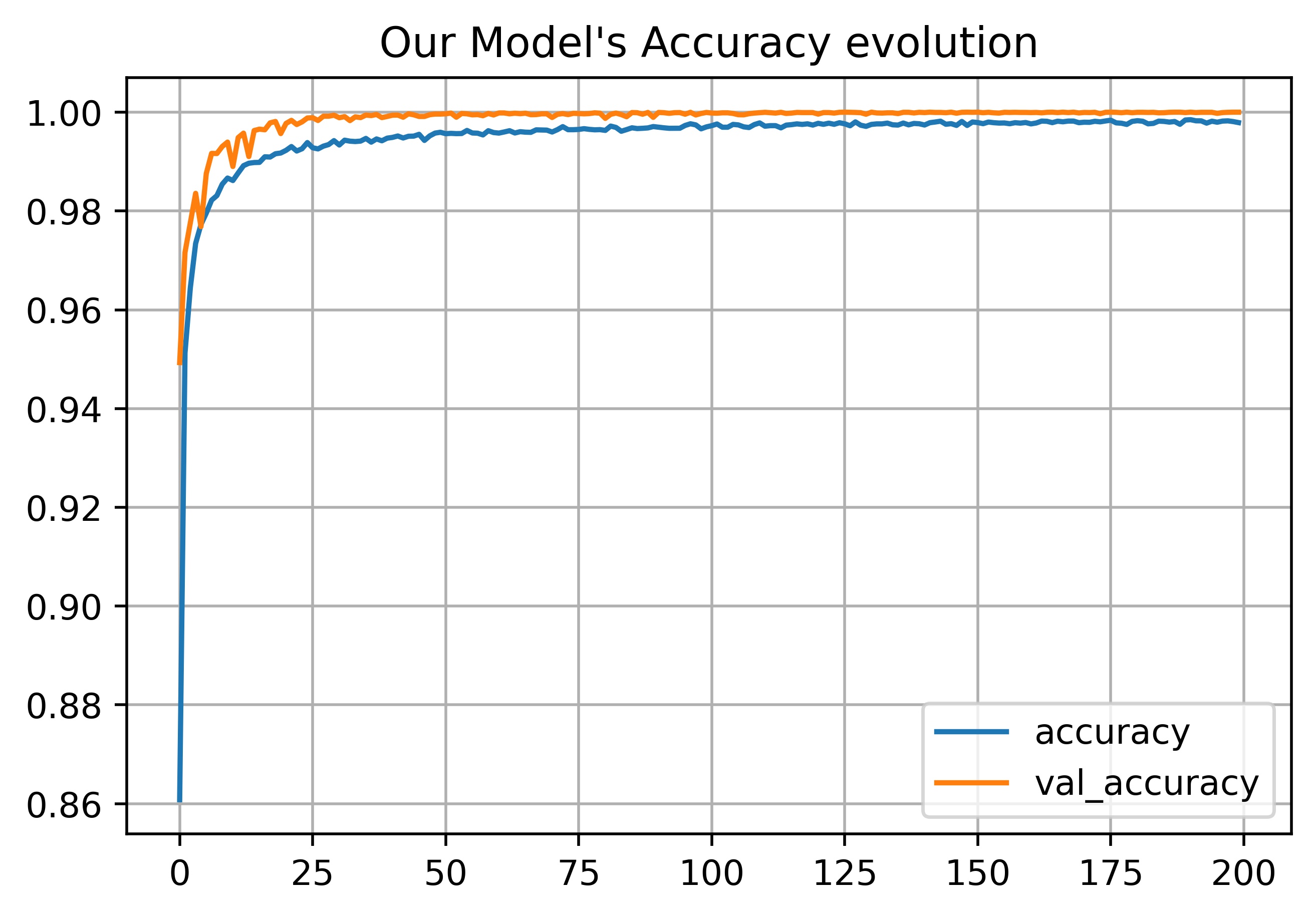}
    }
    \subfloat[Float32\\Confusion Matrix]{
    \includegraphics[width=0.24\linewidth]{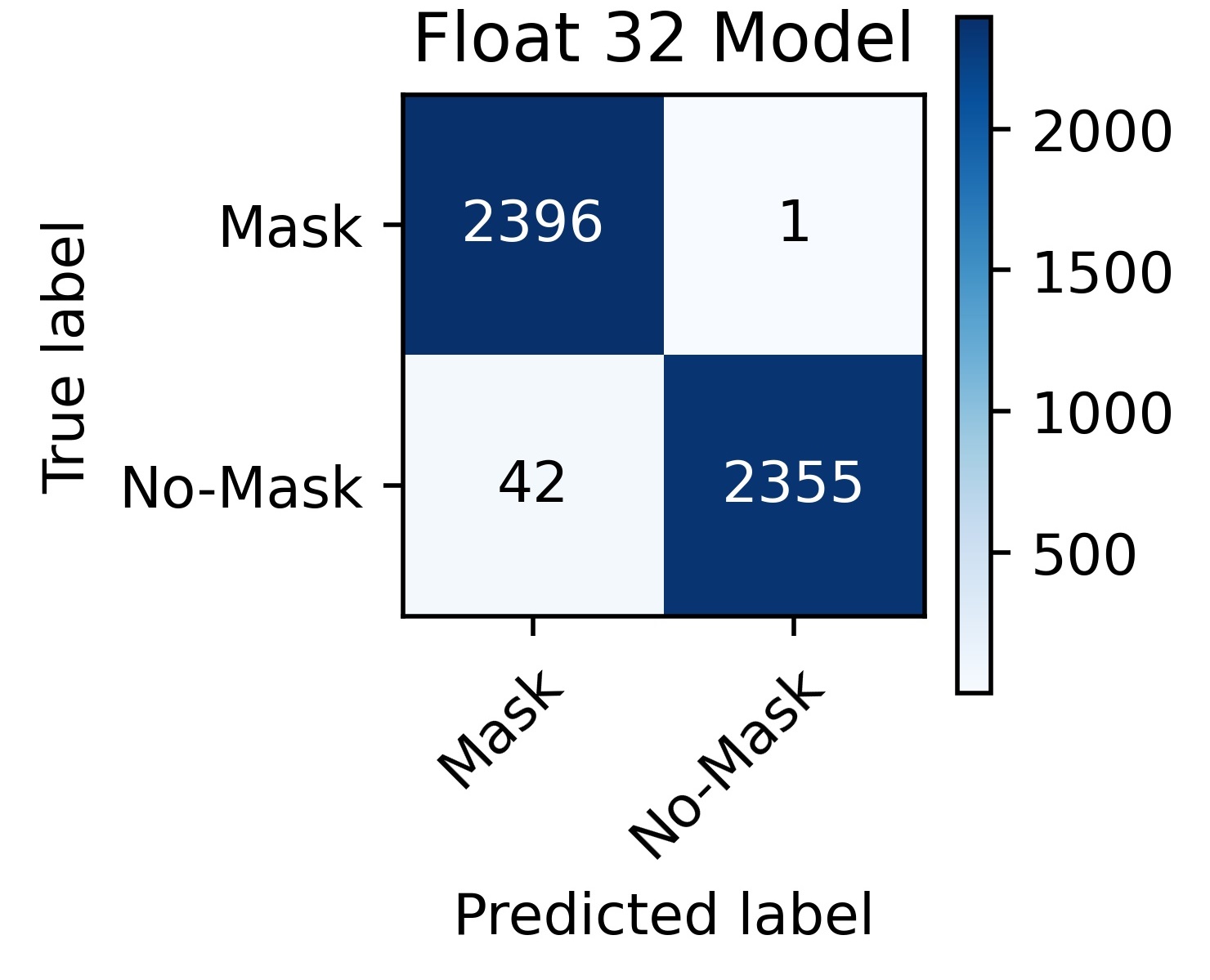}
    }
    \subfloat[Int8\\Confusion Matrix]{
    \includegraphics[width=0.24\linewidth]{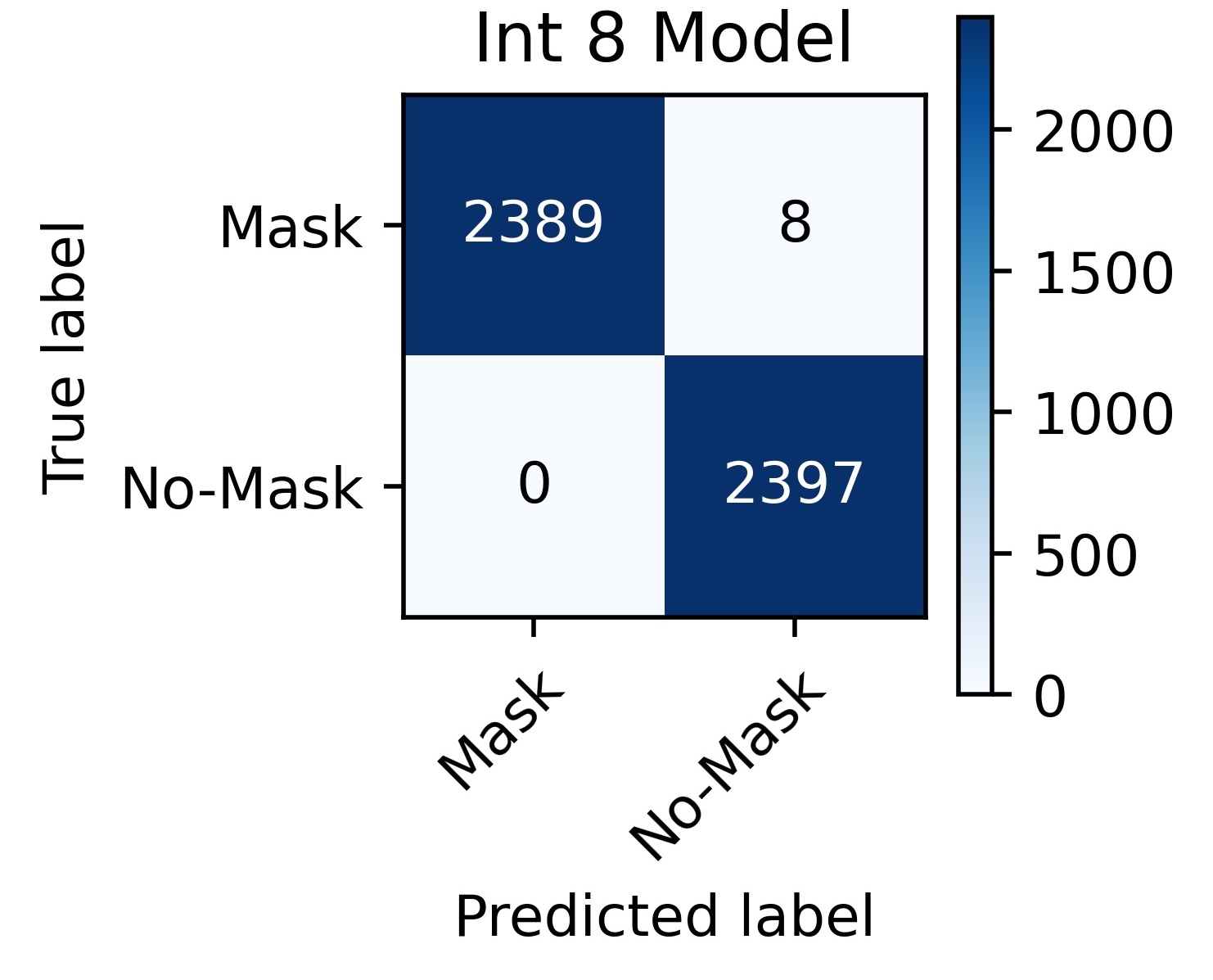}
    }\\
    \vspace{-2mm}
    \caption{Our Model}
    \label{fig:our_acc}

\end{figure}

The proposed model reached the training accuracy of 99.79\%, and achieved a testing accuracy of 99.81\% and 99.83\% for float32 and int8 models respectively. The 1.52 MB float32 model was reduced to 138 KB post-qunatization. Our model outperformed SqueezNet and modified SqueezeNet in both accuracy and size. The details can be seen in Fig. \ref{fig:our_acc}.

\subsection{Discussion}
On comparing the SqueezeNet and the Modified SqueezeNet, it was observed that the modified version, which had two fire modules removed, generalized better than the original model, keeping the precision constant. Thus it was observed,
\vspace{-5mm}
\begin{center}
\textit{“On resource-constrained endpoints, smaller models sometimes outperform bigger ones in generalizing to new data.”}
\end{center}
\vspace{-1mm}

It was also observed,
\vspace{-1mm}
\begin{center}    
\textit{“On devices with Floating Point Unit (FPU) support, keeping inputs and outputs as float32 gives best results.”}
\end{center}
\vspace{-1mm}

The proposed model, despite being the smallest one, achieved the highest accuracy amongst all three. Since the int8 accuracy is a little more than float32 accuracy for all experimental models, the following conclusion can be drawn,
\vspace{-2mm}
\begin{center}
\textit{“Int8 appears to generalize better than float32 for small models.”}
\end{center}
\vspace{-3mm}

\begin{figure}[!h]
%\vspace{2mm}
    \centering
    \includegraphics[width=0.77\linewidth]{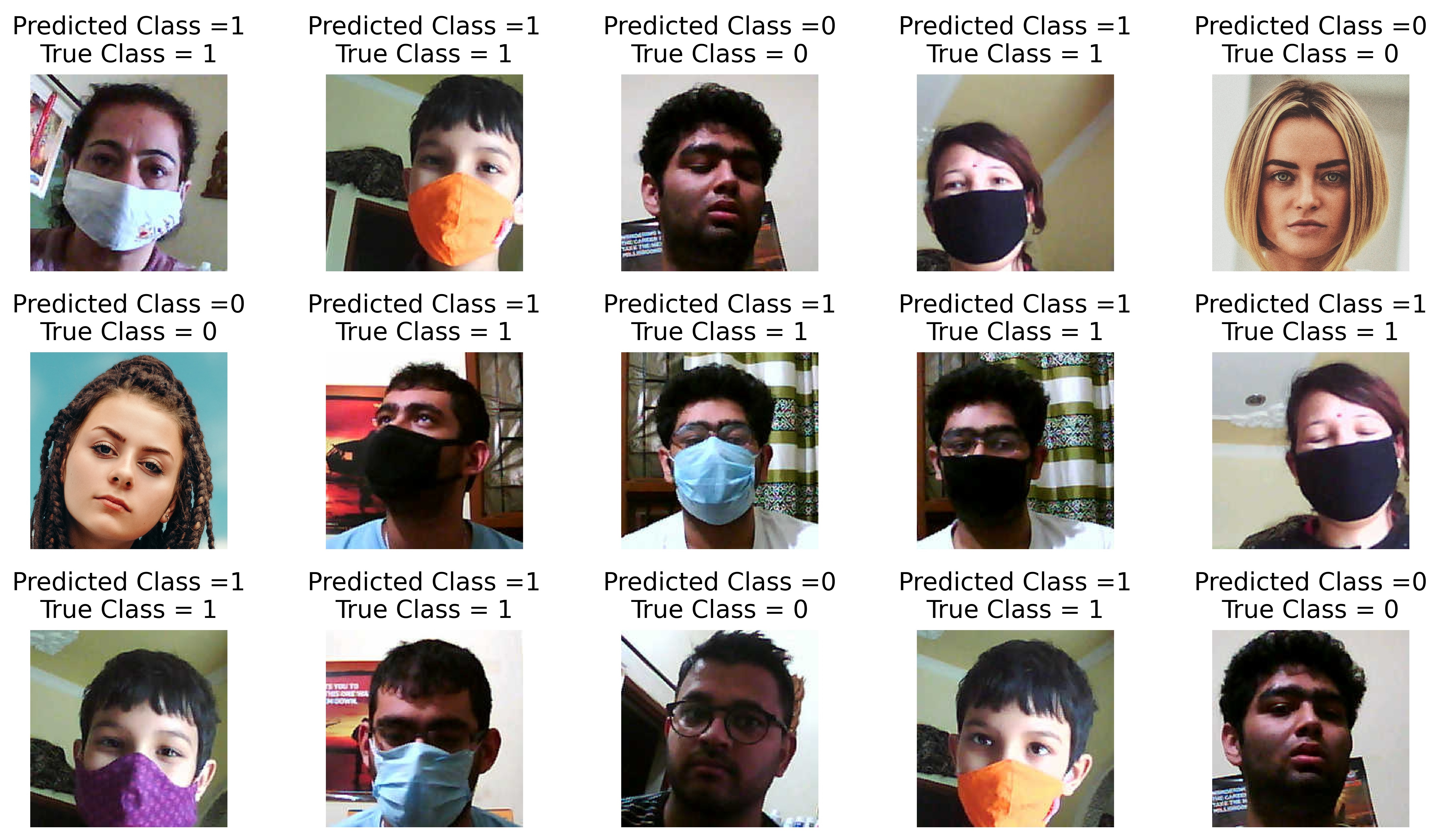}
    \vspace{-3mm}
    \caption{Some predictions made by our int8 model}
    \label{fig:int8images}
    %\vspace{-2mm}
\end{figure}

Table \ref{tab2} shows the size and accuracy comparison of the three models. Table \ref{tabcc} illustrates the compression comparison. The proposed model, despite having the smallest size, achieved the highest accuracy, precision, recall and F1 score, as can be seen in Tables \ref{tab4} and \ref{tab3}, representing the int8 and float32 models respectively. Some of the model predictions can be seen in Fig. \ref{fig:int8images}. Even smaller CNNs may overfit when solving problems like binary classification, hence aggressive regularization is required to increase their generalization accuracy. 

In this research, dropout was used after every layer and it made a significant difference in the test accuracy achieved. Observing the proposed model's architecture in Fig. \ref{fig:model_image} it was found that,
\vspace{-2mm}
\begin{center}
\textit{“Dropout added after every layer seems to significantly improve the generalization of smaller models.”}
\end{center}
\vspace{-2mm}

Interpolation augmentation, as suggested in \cite{ourpaper}, was used in the proposed model, improving generalization, and corroborating the statement,
\vspace{-1mm}
\begin{center}
\textit{“Interpolation Augmentation seems to improve generalization for resource-constrained endpoints.”}
\end{center}
\vspace{-4mm}

\begin{table}[htbp]
        \captionsetup{font=footnotesize}
        \begin{minipage}{0.5\textwidth}
            \centering
            \caption{Model Comparison}
            \label{tab2}
            \scalebox{0.6}{
\begin{tabular}{c c c c c}
\hline
\textbf{Model} & \textbf{\textit{float32}}& \textbf{\textit{float32}}& \textbf{\textit{int8}}& \textbf{\textit{int8}} \\
\textbf{Name} & \textbf{\textit{Size}}& \textbf{\textit{Test Acc.}}& \textbf{\textit{Size}}& \textbf{\textit{Test Acc.}} \\
\hline
SqueezeNet&8.0 MB&98.50 \%&780.0 KB&98.53 \%\\
\hline
Modified Squeezenet&3.8 MB&98.93 \%&386.0 KB&98.99 \%\\
\hline
\textbf{Our Model}&\textbf{1.5 MB}&\textbf{99.81 \%}&\textbf{138.0 KB}&\textbf{99.83 \%}\\
\hline
\end{tabular}}
            
        \end{minipage}
        \hfillx
        \begin{minipage}{0.5\textwidth}
            \centering
            \caption{Compression Comparison}
            \label{tabcc}
            \scalebox{0.6}{
\begin{tabular}{c c@{\quad}c}
\hline
\textbf{Model} & \textbf{\textit{float32} }$\rightarrow$\textbf{\textit{ int8}}& \textbf{\textit{Size Reduction}}\\
\textbf{Name} & \textbf{\textit{Reduction in Size}}& \textbf{\textit{Percentage}}\\
\hline
SqueezeNet&7417 KB&90.48 \%\\
\hline
Modified Squeezenet&3546 KB&90.18 \%\\
\hline
\textbf{Our Model}&\textbf{1428 KB}&\textbf{91.16 \%}\\
\hline
\end{tabular}}
            
        \end{minipage}
        \vspace{2mm}
    \end{table}
\vspace{-1.7mm}
\begin{table}[htbp]
        \captionsetup{font=footnotesize}
        \begin{minipage}{0.5\textwidth}
            \centering
             \vspace{-0.5\baselineskip}
            \caption{int8 Classification Report}
            \label{tab4}
            \scalebox{0.6}{
\begin{tabular}{c c c c }
\hline
\multicolumn{4}{c}{\textbf{SqueezeNet}}\\
\hline
\textbf{Label} & \textbf{\textit{Precision}}& \textbf{\textit{Recall}}& \textbf{\textit{F-1 Score}}\\
\hline
Mask&1&0.97&0.99\\
\hline
No-Mask&0.97&1&0.99\\
\hline
\multicolumn{4}{c}{}\\
\hline
\multicolumn{4}{c}{\textbf{Modified SqueezeNet}}\\
\hline
\textbf{Label} & \textbf{\textit{Precision}}& \textbf{\textit{Recall}}& \textbf{\textit{F-1 Score}}\\
\hline
Mask&1&0.97&0.99\\
\hline
No-Mask&0.97&1&0.99\\
\hline
\multicolumn{4}{c}{}\\
\hline
\multicolumn{4}{c}{\textbf{Our Model}}\\
\hline
\textbf{Label} & \textbf{\textit{Precision}}& \textbf{\textit{Recall}}& \textbf{\textit{F-1 Score}}\\
\hline
Mask&1&1&1\\
\hline
No-Mask&1&1&1\\
\hline
\end{tabular}}
   
        \end{minipage}
        \vspace{-2mm}
        \hfill
        \begin{minipage}{0.5\textwidth}
            \centering
            \vspace{-0.5\baselineskip}
            \caption{float32 Classification Report}
            \label{tab3}
            \scalebox{0.6}{
\begin{tabular}{c c c c}
\hline
\multicolumn{4}{c}{\textbf{SqueezeNet}}\\
\hline
\textbf{Label} & \textbf{\textit{Precision}}& \textbf{\textit{Recall}}& \textbf{\textit{F-1 Score}}\\
\hline
Mask&0.9&1&0.95\\
\hline
No-Mask&1&0.89&0.94\\
\hline
\multicolumn{4}{c}{}\\
\hline
\multicolumn{4}{c}{\textbf{Modified SqueezeNet}}\\
\hline
\textbf{Label} & \textbf{\textit{Precision}}& \textbf{\textit{Recall}}& \textbf{\textit{F-1 Score}}\\
\hline
Mask&0.96&1&0.98\\
\hline
No-Mask&1&0.95&0.98\\
\hline
\multicolumn{4}{c}{}\\
\hline
\multicolumn{4}{c}{\textbf{Our Model}}\\
\hline
\textbf{Label} & \textbf{\textit{Precision}}& \textbf{\textit{Recall}}& \textbf{\textit{F-1 Score}}\\
\hline
Mask&0.98&1&0.99\\
\hline
No-Mask&1&0.98&0.99\\
\hline

\end{tabular}}
            
        \end{minipage}
        \vspace{3.3mm}
    \end{table}

%%%%%%%
\FloatBarrier

\section{Conclusion}
In this research work, an extremely small and well-generalizable CNN based solution has been proposed for face mask recognition on edge devices with extreme resource constraints. The solution has been deployed on a microcontroller development board called OpenMV Cam H7. The model is just 138 KB in size and runs at 30 FPS on the board. It has a test accuracy of 99.83\%.

It has been shown that aggressive regularization through dropout might be useful for developing extremely generalizable CNN architectures for problems like binary classification. The method proposed in this paper is universal and applicable to any microcontroller architecture. The methodology used in this research work can also be used to build and deploy architectures for other challenging problems. The pipeline followed here, which includes, dataset construction, training on float32, quantization to int8, and deployment on edge devices, is applicable to a wide spectrum of resource-constrained, intelligent edge solutions.

Future avenues of research include building systems that are more robust to noise and can work on even smaller microcontrollers. Work can be done on building datasets that include images from more diverse sources. Novel quantization schemes can be developed for converting float32 to int8. Smaller precision can be experimented with, including 6-bit, 4-bit and binarized neural networks too.

\bibliographystyle{spmpsci}
\bibliography{author}

\begin{thebibliography}{10}
\providecommand{\url}[1]{{#1}}
\providecommand{\urlprefix}{URL }
\expandafter\ifx\csname urlstyle\endcsname\relax
  \providecommand{\doi}[1]{DOI~\discretionary{}{}{}#1}\else
  \providecommand{\doi}{DOI~\discretionary{}{}{}\begingroup
  \urlstyle{rm}\Url}\fi

\bibitem{stmicro}
32-bit Arm® Cortex®-M7 480MHz MCUs, up to 2MB Flash, up to 1MB RAM, 46 com.
  and analog interfaces (2019).
\newblock
  \urlprefix\url{https://www.st.com/resource/en/datasheet/stm32h743vi.pdf}

\bibitem{abdelkader2017openmv}
{Abdelkader}, I., {El-Sonbaty}, Y., {El-Habrouk}, M.: {Openmv: A Python
  powered, extensible machine vision camera}.
\newblock arXiv e-prints arXiv:1711.10464 (2017)

\bibitem{banner2019posttraining}
Banner, R., Nahshan, Y., Soudry, D.: Post training 4-bit quantization of
  convolutional networks for rapid-deployment.
\newblock In: H.~Wallach, H.~Larochelle, A.~Beygelzimer, F.~d\textquotesingle
  Alch\'{e}-Buc, E.~Fox, R.~Garnett (eds.) Advances in Neural Information
  Processing Systems, vol.~32, pp. 7950--7958. Curran Associates, Inc. (2019).
\newblock
  \urlprefix\url{https://proceedings.neurips.cc/paper/2019/file/c0a62e133894cdce435bcb4a5df1db2d-Paper.pdf}

\bibitem{chavda2020multistage}
{Chavda}, A., {Dsouza}, J., {Badgujar}, S., {Damani}, A.: {Multi-Stage CNN
  Architecture for Face Mask Detection}.
\newblock arXiv e-prints arXiv:2009.07627 (2020)

\bibitem{choukroun2019lowbit}
{Choukroun}, Y., {Kravchik}, E., {Yang}, F., {Kisilev}, P.: Low-bit
  quantization of neural networks for efficient inference.
\newblock In: 2019 IEEE/CVF International Conference on Computer Vision
  Workshop (ICCVW), pp. 3009--3018 (2019).
\newblock \doi{10.1109/ICCVW.2019.00363}

\bibitem{dong2019hawqv2}
Dong, Z., Yao, Z., Cai, Y., Arfeen, D., Gholami, A., Mahoney, M.W., Keutzer,
  K.: Hawq-v2: Hessian aware trace-weighted quantization of neural networks.
\newblock In: Advances in Neural Information Processing Systems 33
  pre-proceedings (NeurIPS 2020) (2019)

\bibitem{dong2019hawq}
Dong, Z., Yao, Z., Gholami, A., Mahoney, M.W., Keutzer, K.: Hawq: Hessian aware
  quantization of neural networks with mixed-precision.
\newblock In: Proceedings of the IEEE/CVF International Conference on Computer
  Vision (ICCV) (2019)

\bibitem{8019465}
{Ge}, S., {Luo}, Z., {Zhao}, S., {Jin}, X., {Zhang}, X.: Compressing deep
  neural networks for efficient visual inference.
\newblock In: 2017 IEEE International Conference on Multimedia and Expo (ICME),
  pp. 667--672 (2017).
\newblock \doi{10.1109/ICME.2017.8019465}

\bibitem{gong2019differentiable}
Gong, R., Liu, X., Jiang, S., Li, T., Hu, P., Lin, J., Yu, F., Yan, J.:
  Differentiable soft quantization: Bridging full-precision and low-bit neural
  networks.
\newblock In: Proceedings of the IEEE/CVF International Conference on Computer
  Vision (ICCV) (2019)

\bibitem{iandola2016squeezenet}
{Iandola}, F.N., {Han}, S., {Moskewicz}, M.W., {Ashraf}, K., {Dally}, W.J.,
  {Keutzer}, K.: {SqueezeNet: AlexNet-level accuracy with 50x fewer parameters
  and $<$0.5MB model size}.
\newblock arXiv e-prints arXiv:1602.07360 (2016)

\bibitem{jacob2017quantization}
Jacob, B., Kligys, S., Chen, B., Zhu, M., Tang, M., Howard, A., Adam, H.,
  Kalenichenko, D.: Quantization and training of neural networks for efficient
  integer-arithmetic-only inference.
\newblock In: Proceedings of the IEEE Conference on Computer Vision and Pattern
  Recognition (CVPR) (2018)

\bibitem{dataset12k}
Jangra, A.: Face mask ~12k images dataset (2020).
\newblock
  \urlprefix\url{https://www.kaggle.com/ashishjangra27/face-mask-12k-images-dataset}

\bibitem{jiang2020retinamask}
{Jiang}, M., {Fan}, X., {Yan}, H.: {RetinaMask: A Face Mask detector}.
\newblock arXiv e-prints arXiv:2005.03950 (2020)

\bibitem{chowdary2020face}
{Jignesh Chowdary}, G., {Singh Punn}, N., {Sonbhadra}, S.K., {Agarwal}, S.:
  {Face Mask Detection using Transfer Learning of InceptionV3}.
\newblock arXiv e-prints arXiv:2009.08369 (2020)

\bibitem{Lecun98gradient-basedlearning}
LeCun, Y., Bottou, L., Bengio, Y., Haffner, P.: Gradient-based learning applied
  to document recognition.
\newblock In: Intelligent Signal Processing, pp. 306--351. IEEE Press (2001)

\bibitem{99a5a804e1134f46a93a858fb3cb597b}
Loey, M., Manogaran, G., Taha, M., Khalifa, N.: A hybrid deep transfer learning
  model with machine learning methods for face mask detection in the era of the
  covid-19 pandemic.
\newblock Measurement: Journal of the International Measurement Confederation
  \textbf{167} (2021).
\newblock \doi{10.1016/j.measurement.2020.108288}

\bibitem{LOEY2020102600}
Loey, M., Manogaran, G., Taha, M.H.N., Khalifa, N.E.M.: Fighting against
  covid-19: A novel deep learning model based on yolo-v2 with resnet-50 for
  medical face mask detection.
\newblock Sustainable Cities and Society p. 102600 (2020).
\newblock \doi{https://doi.org/10.1016/j.scs.2020.102600}.
\newblock
  \urlprefix\url{http://www.sciencedirect.com/science/article/pii/S2210670720308179}

\bibitem{dataset2}
Makwana, D.: Face mask classification (2020).
\newblock \urlprefix\url{https://www.kaggle.com/dhruvmak/face-mask-detection}

\bibitem{8888092}
{Meenpal}, T., {Balakrishnan}, A., {Verma}, A.: Facial mask detection using
  semantic segmentation.
\newblock In: 2019 4th International Conference on Computing, Communications
  and Security (ICCCS), pp. 1--5 (2019).
\newblock \doi{10.1109/CCCS.2019.8888092}

\bibitem{nahshan2020loss}
{Nahshan}, Y., {Chmiel}, B., {Baskin}, C., {Zheltonozhskii}, E., {Banner}, R.,
  {Bronstein}, A.M., {Mendelson}, A.: {Loss Aware Post-training Quantization}.
\newblock arXiv e-prints arXiv:1911.07190 (2019)

\bibitem{covidEnc}
Paul, A.J.: Recent advances in selective image encryption and its
  indispensability due to covid-19.
\newblock In: IEEE Recent Advances in Intelligent Computational Systems
  (RAICS), 2020. (in press)

\bibitem{ourpaper}
Paul, A.J., Mohan, P., Sehgal, S.: Rethinking generalization in american sign
  language prediction for edge devices with extremely low memory footprint.
\newblock In: IEEE Recent Advances in Intelligent Computational Systems
  (RAICS), 2020. (in press)

\bibitem{Roy2020}
Roy, B., Nandy, S., Ghosh, D., Dutta, D., Biswas, P., Das, T.: Moxa: A deep
  learning based unmanned approach for real-time monitoring of people wearing
  medical masks.
\newblock Transactions of the Indian National Academy of Engineering
  \textbf{5}(3), 509--518 (2020).
\newblock \doi{10.1007/s41403-020-00157-z}.
\newblock \urlprefix\url{https://doi.org/10.1007/s41403-020-00157-z}

\bibitem{yao2020hawqv3}
{Yao}, Z., {Dong}, Z., {Zheng}, Z., {Gholami}, A., {Yu}, J., {Tan}, E., {Wang},
  L., {Huang}, Q., {Wang}, Y., {Mahoney}, M.W., {Keutzer}, K.: {HAWQV3: Dyadic
  Neural Network Quantization}.
\newblock arXiv e-prints arXiv:2011.10680 (2020)

\bibitem{zhao2019improving}
Zhao, R., Hu, Y., Dotzel, J., De~Sa, C., Zhang, Z.: Improving neural network
  quantization without retraining using outlier channel splitting.
\newblock In: K.~Chaudhuri, R.~Salakhutdinov (eds.) Proceedings of the 36th
  International Conference on Machine Learning, \emph{Proceedings of Machine
  Learning Research}, vol.~97, pp. 7543--7552. PMLR, Long Beach, California,
  USA (2019).
\newblock \urlprefix\url{http://proceedings.mlr.press/v97/zhao19c.html}

\end{thebibliography}
\end{document}